\documentclass[11pt]{article}
\usepackage{coling2020}
\usepackage{times}
\usepackage{url}
\usepackage{latexsym}
\usepackage{graphicx}
\usepackage{subfigure}

\colingfinalcopy 

\title{Multi-label classification of promotions in digital leaflets \\ using textual and visual information}

\author{Roberto Arroyo, David Jim\'enez-Cabello and Javier Mart\'inez-Cebri\'an \\
  Nielsen Connect R\&D AI \\
  \small{Calle Salvador de Madariaga, 1, 28027, Madrid, Spain} \\
  \small{https://www.nielsen.com/} \\
  \small{\{roberto.arroyo, david.jimenez, javier.martinezcebrian\}@nielsen.com} \\}

\date{}

\begin{document}
\maketitle
\begin{abstract}
    Product descriptions in e-commerce platforms contain detailed and valuable information about retailers assortment. In particular, coding promotions within digital leaflets are of great interest in e-commerce as they capture the attention of consumers by showing regular promotions for different products. However, this information is embedded into images, making it difficult to extract and process for downstream tasks. In this paper, we present an \mbox{end-to-end} approach that classifies promotions within digital leaflets into their corresponding product categories using both visual and textual information. Our approach can be divided into three key components: 1)~region detection, 2)~text recognition and 3)~text classification. In many cases, a single promotion refers to multiple product categories, so we introduce a multi-label objective in the classification head. We demonstrate the effectiveness of our approach for two separated tasks: 1) \mbox{image-based} detection of the descriptions for each individual promotion and 2) \mbox{multi-label} classification of the product categories using the text from the product descriptions. We train and evaluate our models using a private dataset composed of images from digital leaflets obtained by Nielsen. Results show that we consistently outperform the proposed baseline by a large margin in all the experiments.
\end{abstract}

\section{Introduction}
\label{sec:introduction}
\newcommand{\ie}{\textit{i}.\textit{e}. }
\newcommand{\eg}{\textit{e}.\textit{g}. }

\begin{figure}[!h]
\centering
\subfigure[Region-based detection.]                                     
{\includegraphics[width=0.32\textwidth]{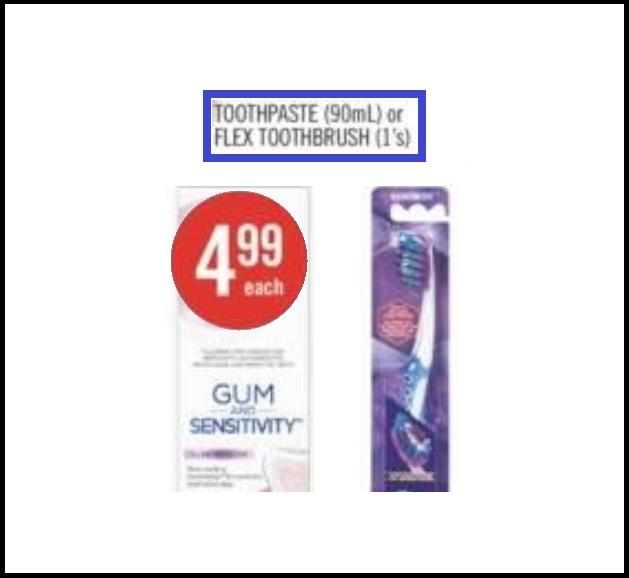}}
\subfigure[Text recognition and extraction.]                                     
{\includegraphics[width=0.32\textwidth]{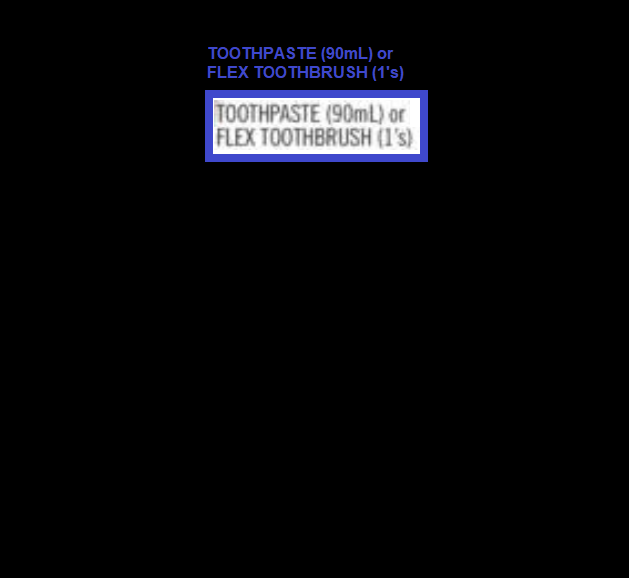}}
\subfigure[Multi-label text classification.]                                     
{\includegraphics[width=0.32\textwidth]{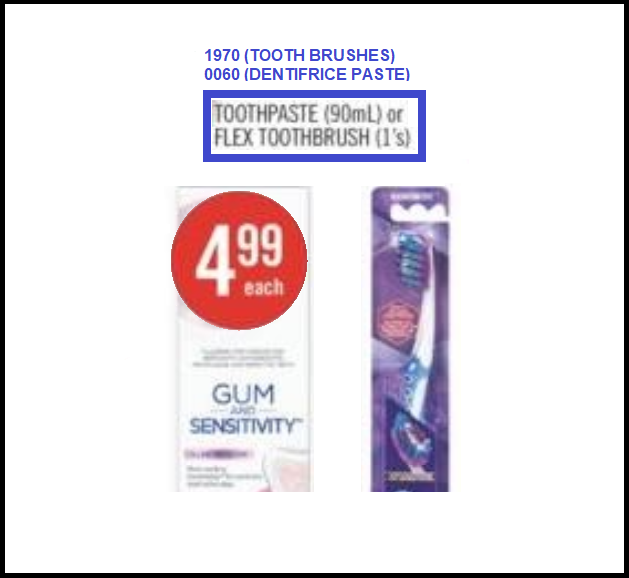}}
\vspace{-2mm}
\caption{Key components of our approach for a single promotion within a digital leaflet.}
\vspace{-2mm}
\label{fig:leaflets_categorization_parts} 
\end{figure}

The latest advances in Artificial Intelligence (AI) have provided new tools to enhance the automation of different recognition problems. We are witnessing a clear trend to merge different domains within AI to obtain better representations for the most complex problems. Many recent approaches merge textual and visual information by applying Natural~Language~Processing (NLP) and Computer~Vision~(CV), with the aim of solving problems that involve both text and images~\cite{ref:Bai18ieeea}. Within this context, structured knowledge extraction from unstructured text is an open problem in the e-commerce literature~\cite{ref:Arroyo19wcvpr}. Regardless the source of the information (\eg product websites, product images captured from stores or digital leaflets), it refers to a unified concept that can be denoted as ``automated product coding'', \ie the extraction of attribute values of e-commerce products (see Fig.~\ref{fig:leaflets_categorization_parts}).

Our present work focuses on the case of knowledge extraction from digital leaflets. Most retailers are replacing physical leaflets, that are directly collected from the stores, with digital leaflets that are uploaded in the cloud on the retailers websites. Comparing to real-world e-commerce platforms that contain billions of products with detailed descriptions (including ratings and opinions), digital leaflets include concise textual and visual information of promotions that applies to some of the products of the store assortment for a short period of time and thus it has to be updated regularly. The knowledge extraction of digital leaflets is of great interest for the e-commerce business as it impacts not only on several aspects of the consumers behaviour or seasonality, but also modifies relevant attributes of the products periodically, \eg price or volume.

In this paper, we bring for the first time the problem of automated product coding in digital leaflets for e-commerce. In particular, we present an approach to predict the product categories for each of the promotions within the leaflet that serves as a strong baseline for future works. Technically, this is a multi-label text classification problem as some promotions can potentially apply to several product categories. For that purpose, we hypothesize that most of the information of the promotions is self-contained in the products descriptions, so we first detect all these regions within the image that contain textual descriptions and extract the text using Optical Character Recognition (OCR) techniques.
In Fig.~\ref{fig:leaflets_categorization_parts}, we show a visual representation of the three key components in the proposed approach: 1) region-based detection of the promotions description, 2) text recognition and extraction, and 3) multi-label text classification.

\vspace{0.3cm}
The building blocks depicted in Fig.~\ref{fig:leaflets_categorization_parts} also relate to the different domains covered in the approach and they can be divided into the following three categories: 1) \textbf{CV}: a region detection architecture based on deep learning and image processing to detect the descriptions of each individual promotion within the digital leaflet, 2) \textbf{CV+NLP}: a text recognition method for extracting the textual information contained into the detected descriptions based on OCR and 3) \textbf{NLP}: a multi-label text classification model based on sub-word text embeddings and a shallow neural network.

The main contributions of the paper are three fold:
\begin{enumerate}
    \item We bring for the first time the problem of automated item coding in digital leaflets for e-commerce platforms.
    \item We formulate the process for predicting the categorization of each individual promotion within digital leaflets as a multi-label classification problem, which uses both CV and NLP techniques for properly fusing image and text information.
    \item We conduct several experiments to assess the performance of the model for several aspects: a)~detection of the product description region in promotions, b)~multi-label classification of the product categories and c)~multi-lingual capabilities.
\end{enumerate}

The contents of the paper are structured as follows: related works are described in Section~\ref{sec:related_work}. The technical proposal presented in this paper is described in Section~\ref{sec:our_proposal}. The data used in the evaluations of our proposal, the experiments carried out to validate it and several comparative results are reviewed in Section~\ref{sec:experiments_and_results}. The final conclusions derived from this paper are discussed in Section~\ref{sec:conclusions}.

\section{Related Work}
\label{sec:related_work}

Prior works related to the fusion of NLP and CV have experienced an important growth in the last years due to the advances in deep learning and its influence in both domains. It covers several fields such as text retrieval~\cite{ref:Gomez18eccv}, image detection and classification~\cite{ref:Bai18ieeea} or automated item coding~\cite{ref:Arroyo19wcvpr}. 

Regarding the first step of the system proposed in this paper, approaches based on region detection over images have received an incredible attention in the last few years. The popularization of deep learning jointly with Convolutional Neural Networks (CNN)~\cite{ref:Krizhevsky12nips} has completely changed the traditional paradigm in CV. Standard CNNs are commonly applied only for image classification. However, \mbox{R-CNNs} (R stands for Region-based) are focused on object detection, which combines both detection and classification. Nowadays, techniques such as \mbox{Faster~R-CNN}~\cite{ref:Ren15nips} are broadly extended to localize and classify objects over images. In this method, the regions in the \mbox{R-CNN} are detected by a selective search algorithm based on a Region~Proposal~Network~(RPN). YOLO~\cite{ref:Redmon16cvpr} is also a technique very popularized for object detection which is focused on Single~Shot~Detection~(SSD)~\cite{ref:Liu16eccv}. Similar proposals based on \mbox{R-CNN} architectures can be used in our approach to initially detect the regions where the text of the leaflets descriptions is located over the images.
  
Following with the second stage for the recognition of the texts contained in descriptions regions, Optical Character recognition (OCR) is a broad topic covered in the AI community and it aims at extracting text from images, thus working in the intersection between CV and NLP. The most recent approaches use deep learning techniques~\cite{ref:Lee16cvpr} to examine images pixel by pixel, looking for shapes that match the character traits. Available OCR engines comprise implemented solutions that are \mbox{open-source} and proprietary. Calamari~\cite{ref:Wick20dhq} or Tesseract~\cite{ref:Zacharias20corr} are some of the most effective open-source approaches, with lots of users around the world. However, proprietary solutions such as Google~OCR\footnote{https://cloud.google.com/vision/docs/ocr} are currently obtaining better results in text recognition, including support for a larger number of languages. Our goal is to apply OCR-based algorithms over the regions previously detected using a R-CNN architecture in order to obtain the product descriptions over the images of leaflets.

In the final stage of our described approach, the textual information extracted from the detected regions is classified into their corresponding product categories. The state of the art in short text classification is recently moving to approaches based on DNNs~\mbox{(Deep Neural Networks)}. On the one hand, in~\cite{ref:Joulin17eacl} the authors proposed to incorporate sub-word level information to train textual embeddings very efficiently for text classification. On the other hand, the BERT architecture described in~\cite{ref:Devlin18naacl} also supposed a great milestone in natural language modeling introducing a self-supervised learning strategy that is able to incorporate an architecture based on Transformers~\cite{ref:Vaswani17nips} leveraging a large corpus for training. The embeddings obtained using BERT approaches highly correlate with the linguistic context within a sentence. Thus, proposals based on sub-word level information are very competitive compared to BERT models on those cases where we have unstructured textual information and probably OCR errors, such as the descriptions processed in most of the leaflets. 
It must be also noted that although BERT is focused on standard text processing, there are derived approaches that are also diving into text processing associated with images, such as ViLBERT~\cite{ref:Lu19nips} or \mbox{VL-BERT}~\cite{ref:Su20iclr}. The difference is that these recent approaches are not directly applied to classification over text contained in images, they are used for tasks that involve images and related external text, such as VQA~(Visual~Question~Answering)~\cite{ref:Antol15iccv}.

\section{Our Proposal for Digital Leaflets Categorization}
\label{sec:our_proposal}

The leaflets categorization proposal presented in this paper is focused on the prediction of multiple product categories from images such as the depicted in Fig.~\ref{fig:leaflets_categorization_parts}, which is showing a part of a catalog representing specific products. The solution can be divided into the following three main parts:

\begin{enumerate}
 \item Detection of the regions related to the textual descriptions of each product in a promotion.
 \item Recognition of the associated text inside the regions of the detected descriptions.
 \item Classification of the recognized text into the different product categories of interest.
\end{enumerate}

In this section, we introduce these three main components of the proposed approach, jointly with the whole leaflets categorization pipeline that combines them to obtain the final output.

\subsection{Region-based Detection}

\begin{figure} [!ht] \includegraphics[width=\textwidth]{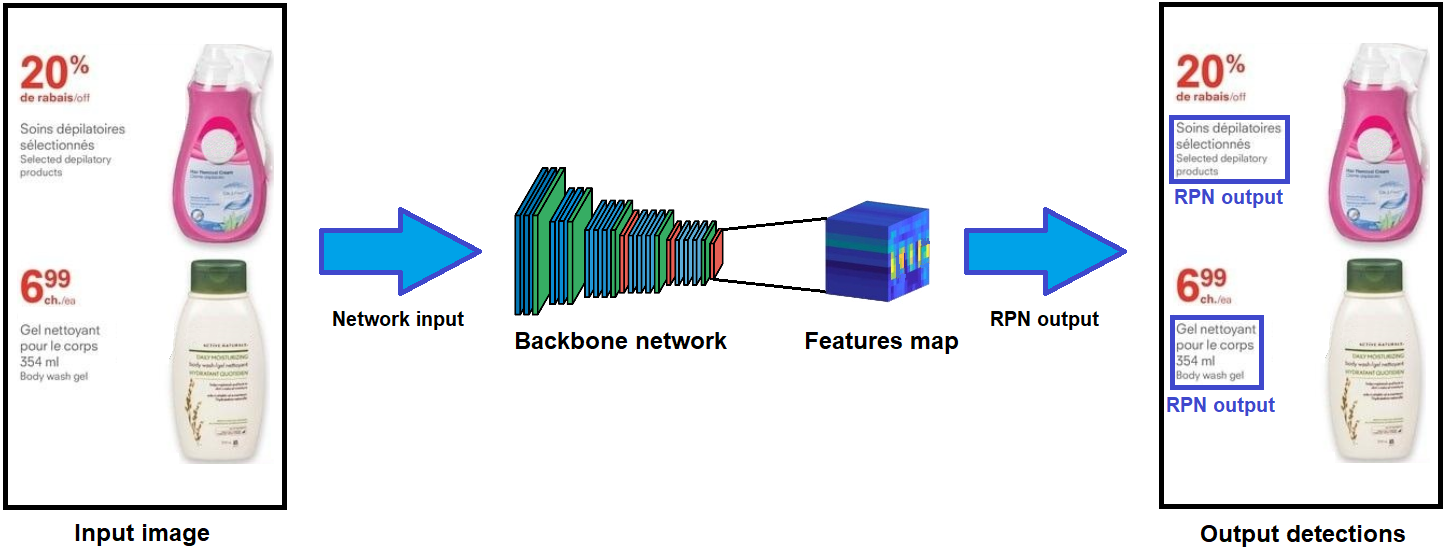}
\caption{Visual representation of the text region detection over images in our \mbox{R-CNN} architecture.} 
\label{fig:RPN_example} 
\end{figure}

The method designed for detecting the regions that contain the texts associated with product descriptions is based on an \mbox{R-CNN} schema, as presented in Fig~\ref{fig:RPN_example}. We decided to use this CV approach because the texts of product descriptions over the images have a specific appearance format that can be effectively visually differentiated, even when several templates and styles are used for varied retailers. Then, the texts from descriptions can be effectively separated from the rest of the texts in the image.

\vspace{0.3cm}
Initially, we tried to differentiate the product descriptions from the rest of the texts in the digital leaflets by only considering predictions with high confidence in the \mbox{multi-label} text classification proposed for the third part of the system (described in detail in Section~\ref{sec:text_classification}). Unfortunately, a great amount of texts out of scope were classified with high confidences by the model, so we decided to design the approach based on the R-CNN architecture to obtain more accurate results in the overall process.

As can be seen in Fig.~\ref{fig:RPN_example}, our architecture applies an internal Region~Proposal~Network~(RPN)~\cite{ref:Ren15nips} with the aim of detecting the positions of the regions of interest. First of all, the image is resized before feeding it into the backbone CNN. This resizing is important to have similar detection schemas independently of the different sizes that could have the input images. For every point in the output, the network has to learn whether a text description region is present in the image at its corresponding position and estimate its size. Several anchors over the input image are used for each location from the backbone network. These anchors indicate possible objects in various sizes and aspect ratios at this location. As the RPN walks through each pixel in the feature map, it has to validate whether these corresponding anchors spanning the input image contain regions of interest. Besides, it has to refine the coordinates of anchors to provide bounding boxes as proposed regions associated with the different text of products descriptions. In order to help with this process, Non-Maximum~Suppression~(NMS)~\cite{ref:Rothe14accv} is applied as follows:

\begin{enumerate}
 \item Choose the bounding box that has the highest confidence score.
 \item Compute its overlap with the rest of bounding boxes and remove the bounding boxes that overlap more than an Intersection~over~Union~(IoU)~\cite{ref:Rezatofighi19cvpr} threshold.
 \item Return to the first step and iterate until there are no more boxes with a lower confidence score than the chosen box.
\end{enumerate}

In order to train the detection model, our architecture requires Ground-Truth (GT) information about bounding boxes from sample images, with the aim of training the network to localize the regions of interest. In standard R-CNN architectures, a part of the network is in charge of classifying the bounding boxes into several classes. However, in our schema this is not required, because we do not need to differentiate the class of the bounding boxes detected, what we need is to classify the internal textual information in the stage of text classification that is detailed in Section~\ref{sec:text_classification}. Then, the standard network part for classifying the obtained visual embeddings is not applied in our model, only the part for localization based on RPN previously explained.

\subsection{Text Recognition and Extraction}

A method based on OCR is used in this second stage in order to recognize the text associated with the previously detected product descriptions. We use Google OCR as basis of our text recognition pipeline. 
Besides, the goal of our work is not focused on contributing a new complete OCR engine, which is a research out of the scope of this paper that considers a full detection, recognition and classification schema for leaflets categorization. 

In our case, OCR converts leaflets images into \mbox{machine-readable} text data. The human visual system reads text by recognizing the patterns of light and dark, translating those patterns into characters and words, and then attaching meaning to it. Similarly, OCR attempts to mimic our visual system by using neural networks.

The approach applied in this stage to compute OCR returns the characters, words and paragraphs obtained from images and their locations. Initially, we implemented the idea of directly clustering the words recognized inside a bounding box detected for a product description, with the aim of providing the whole text string associated with that specific product description. However, we observed that the resulting text string sometimes contained errors due to other \mbox{out-of-scope} texts around the text of interest that interfere with it. To minimize the impact of this recognition issue, we decided to apply a mask to blacken all the parts of the image that are not contained inside the bounding boxes detected in the previous stage by the region detection model. Then, the blackened regions do not interfere with the regions of interest related to product descriptions during the OCR computation. The described blackened process is exemplified in Fig.~\ref{fig:leaflets_categorization_parts}~(b).

Finally, the text extracted by the OCR is post-processed to reduce typical errors, such as the ones associated with strange symbols incorrectly detected, problems derived from lower and upper case letters or dictionary-based corrections.

\subsection{Multi-label Text Classification}
\label{sec:text_classification}

After recognizing the text corresponding to product descriptions in digital leaflets images, a text classification model is applied to predict the different product categories of interest. Each product can be associated with more than one category, so this use case can be considered as an instance \mbox{multi-label} classification problem.

\begin{figure} [!ht] \centering \includegraphics[width=0.95\textwidth]{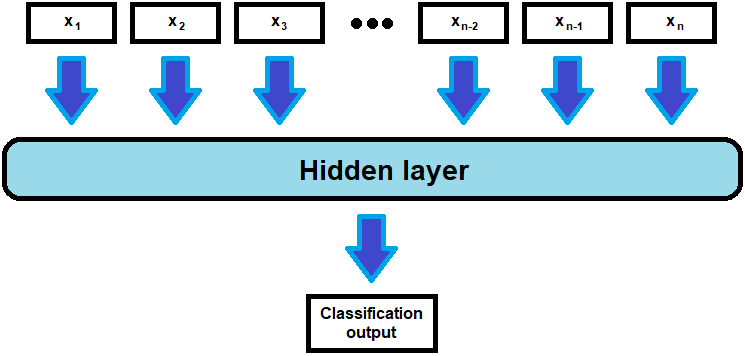}
\caption{FastText architecture used for our multi-label text classification with n-gram features $x_1, x_2, x_3, ... , x_{n-2}, x_{n-1}, x_n$.} 
\label{fig:text_classification_architecture} 
\end{figure}

The proposed text classification model is based on FastText \cite{ref:Joulin17eacl}, as it efficiently scales in the number of categories to predict. The defined architecture is a simple neural network that contains only one layer. The architecture generates a \mbox{bag-of-words} representation of the text, where the embeddings are fetched for every single word. After that, the embeddings are averaged to obtain a single embedding for the whole text in the hidden layer. Once the averaged embeddings are computed, the single vector is fed to independent binary classifiers for each label \mbox{(one-vs-all loss)}. Character \mbox{n-grams} are used, which are really beneficial for text classification problems based on product descriptions (not natural language as it is usually known) and that may also include typos from the OCR. In order to visually understand the architecture and n-grams computation, Fig.~\ref{fig:text_classification_architecture} is presented.

For training the text classification model, a dataset with manually labeled annotations about the categories associated with each text description is required, as explained in detail in Section~\ref{sec:leaflets_dataset}. The trained model is used to perform the inference of the categories related to each promotion description. The inference output gives a vector with probabilities for each available category. A threshold is used to filter the categories corresponding to an instance based on the obtained probabilities, with the aim of providing the \mbox{multi-label} classification. The categories with probabilities above this threshold are considered as positive.

\subsection{Whole Leaflets Categorization Pipeline}

\begin{figure} [!ht] \includegraphics[width=\textwidth]{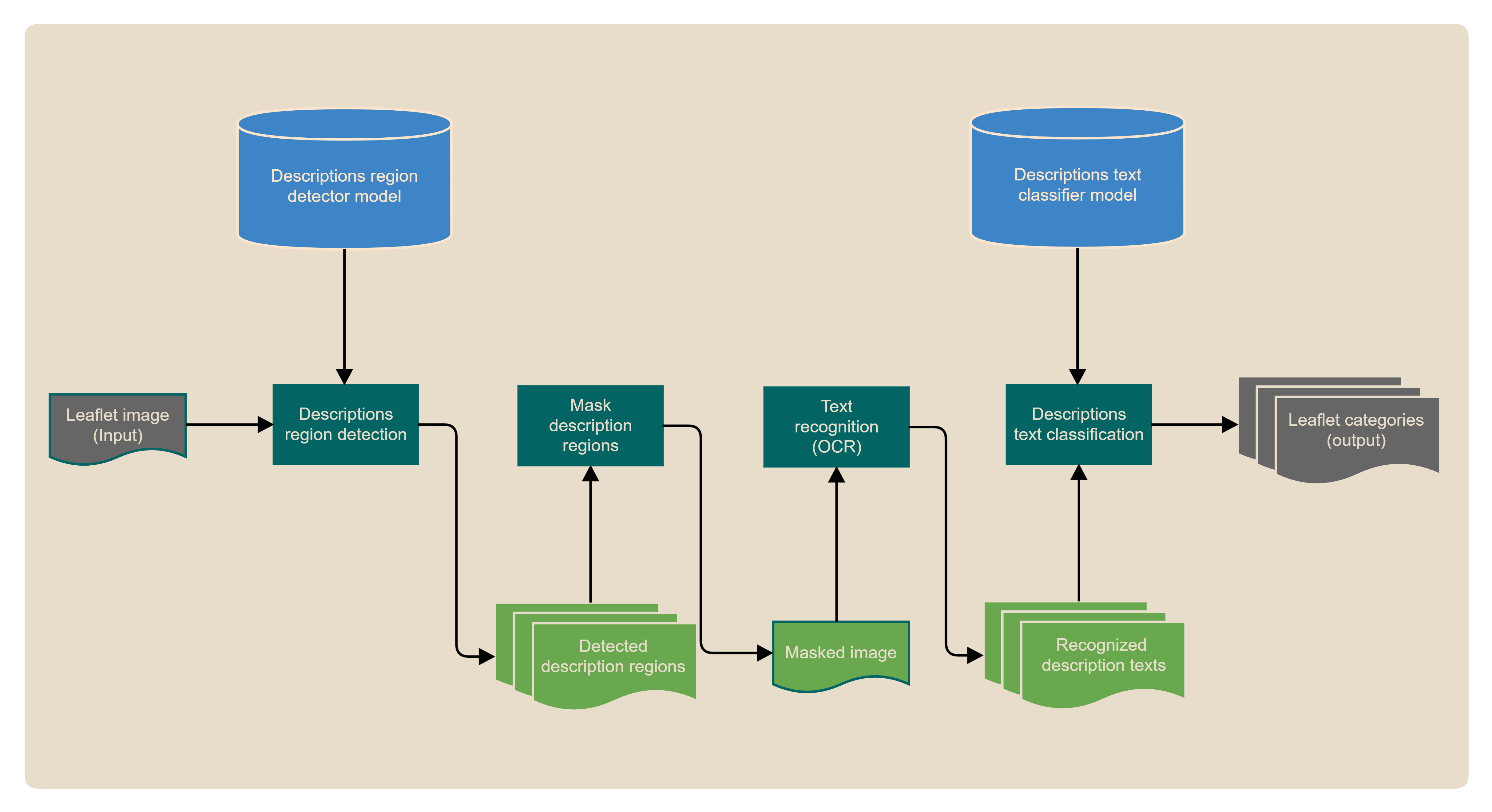}
\caption{Diagram of the whole leaflets categorization pipeline.} 
\label{fig:leaflets_categorization_pipeline} 
\end{figure}

The diagram presented in Fig.~\ref{fig:leaflets_categorization_pipeline} illustrates the overall setup of the solution for the whole leaflets categorization pipeline. An image corresponding to a digital leaflet is received as the input of the system. In the first step, the descriptions related to products are initially identified over the leaflet image using the previously trained region detection model. The detected regions are used for generating the masked image that is used in the text recognition stage to extract the texts of interest for each promotion. Finally, the model for text classification is applied for each description in order to compute the final output, which contains the categories associated for each promotion.

\section{Experiments and Results}
\label{sec:experiments_and_results}

With the aim of validating the proposed approach in a specific leaflets categorization use case, we prepared a set of experiments with leaflets data captured by Nielsen. In this section, we describe the experimental setup: the datasets, the hyperparameters used for training and the comparative results with different data and approaches.

\subsection{Leaflets Dataset}
\label{sec:leaflets_dataset}

\begin{figure}[!ht]
\centering
\subfigure[Base dataset.]                                     
{\includegraphics[width=0.49\textwidth]{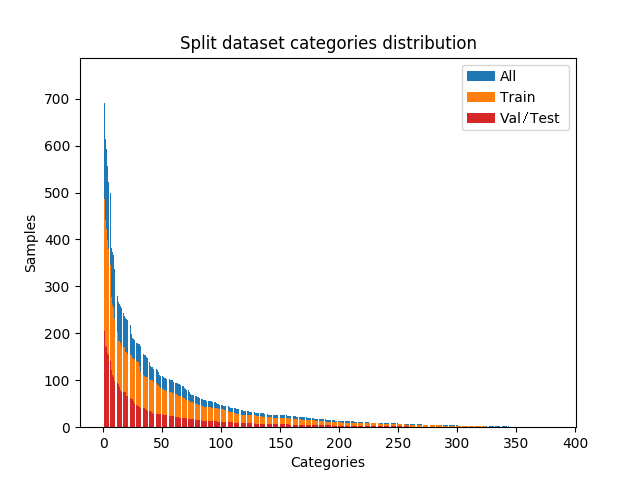}}
\subfigure[Extended dataset.]                                     
{\includegraphics[width=0.49\textwidth]{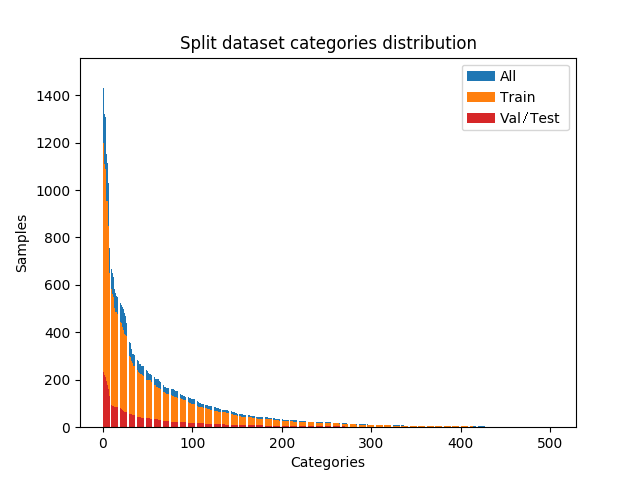}}
\caption{Distribution of category samples for the used leaflets datasets.}
\label{fig:datasets_distribution} 
\end{figure}

To the best of our knowledge, there are not public datasets with GT information for leaflets categorization over images of catalogs, so we used our own labeled datasets from Nielsen internal data. We applied two different leaflets datasets for training and evaluation. Firstly, a "base" dataset with leaflets from only one retailer with textual descriptions in English. Secondly, a extended dataset with leaflets from four retailers with varied image formats to test generalization, and texts from two languages (English and French) to evaluate the multi-lingual capabilities of our approach. As the datasets are composed of proprietary images, we can not publicly share them. However, the main properties and statistics from both datasets are summarized in Table~\ref{table:dataset_statistics}. It must be noted that the data distribution is long-tailed and thus unbalanced for both training and validation/test splits, as shown in Fig.~\ref{fig:datasets_distribution}. Then, this is an extra challenge for our models in order to be robust against the typical problems derived from \mbox{long-tail} datasets.
\vspace{0.3cm}
\begin{table}[!ht]
\begin{center}
\small
\begin{tabular}{|l|c|c|c|c|c|c|c|}
\hline
\multicolumn{1}{|c|}{\textbf{Dataset}} & \textbf{\#Languages} & \textbf{\#Retailers} & \textbf{\#Images} & \textbf{\#Samples} & \textbf{\#Categories} & \textbf{\begin{tabular}[c]{@{}c@{}}Avg. samples \\ per cat.\end{tabular}} & \textbf{\begin{tabular}[c]{@{}c@{}}Std. samples\\ per cat.\end{tabular}} \\ \hline
\textbf{Base}                          & 1                    & 1                    & 449               & 10,333             & 382                   & 27.05                                                                     & 100.75                                                                   \\ \hline
\textbf{Extended}                      & 2                    & 4                    & 1,079             & 20,646             & 504                   & 40.96                                                                     & 189.79                                                                   \\ \hline
\end{tabular}
\end{center}
\caption{Main statistics about the leaflets categorization datasets used for training and evaluation.}
\label{table:dataset_statistics}
\end{table}

\subsection{Hyperparameters Tuning}

Region detection and text classification models require some hyperparameters tuning to obtain the best possible results. Some standard hyperparameters typically used in these models are configured.

In region detection, we decided to train our models using pre-trained weights on ImageNet~\cite{ref:Deng09cvpr} and based on a ResNet-101 backbone~\cite{ref:He16cvpr}. Anchor scales and ratios for RPN are important hyperparameters, which are configured as $[2, 4, 8]$ and $[0.5, 1, 2]$, respectively. Learning rate was set to $1\cdot10^{-6}$ and regularization is applied by means of dropout (keep probabilities mode), which is set to $0.7$. Besides, an Adam optimizer~\cite{ref:Kingma15iclr} is used. A confidence threshold of $0.4$ is applied to discard bounding boxes with low confidences. Trainings are iterated during $100$ epochs.

The \mbox{multi-label} text classification hyperparameters configuration has a great dependence on the number of \mbox{n-grams}, which is a value finally set to $3$ based on previous \mbox{cross-validation} experiments. Besides, the learning rate is set to $0.1$ with a learning update rate of $100$. A confidence threshold of $0.25$ is applied to identify the categories of interest for a specific product description. We train for $30$ epochs.

\subsection{Results in Leaflets Categorization}

In order to evaluate the performance of our approach, we apply metrics based on precision, recall and accuracy. Besides, we also use these metrics to obtain comparative results with respect to a standard baseline text classification, which is based on directly extracting OCR paragraphs on the wild from the image, without previously using a RPN detector for filtering texts related to product descriptions. In this baseline case, the texts recognized by the OCR with all the class probabilities below the text classifier confidence threshold are not considered as descriptions. Overall test results comparison is presented in Table~\ref{table:results_base} for the base dataset. As can be seen in these results, our approach yields an accuracy improvement of 24 points with respect to the standard baseline. These results confirm the enhancement given by our system with respect to the proposed baseline.

\begin{table}[!ht]
\begin{center}
\begin{tabular}{|l|c|c|c|}
\hline
\multicolumn{1}{|c|}{\textbf{Method}}                     & \textbf{Precision} & \textbf{Recall} & \textbf{Accuracy} \\ \hline
Baseline (OCR on the wild + text classification) & 0.64               & 0.66            & 0.48              \\ \hline
\textbf{Ours (RPN + OCR masked + text classification)}    & \textbf{0.86}               & \textbf{0.81}            & \textbf{0.72}              \\ \hline
\end{tabular}
\end{center}
\caption{Overall test results comparing a standard baseline approach vs ours in the base leaflets dataset.}
\label{table:results_base}
\end{table}

\begin{figure}[!ht]
\centering
\subfigure[Baseline (OCR on the wild + text classification).]                                     
{\includegraphics[width=0.45\textwidth]{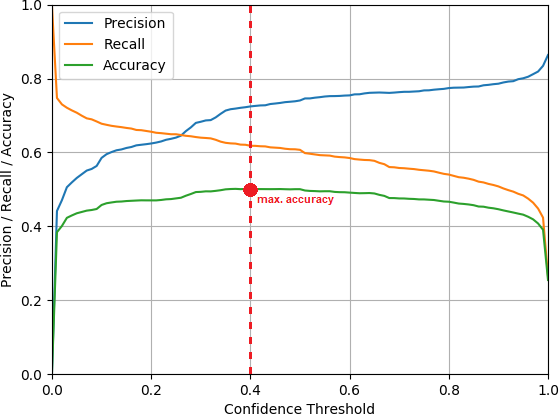}} \hspace{0.05\textwidth}
\subfigure[Ours (RPN + OCR masked + text classification).]                                     
{\includegraphics[width=0.45\textwidth]{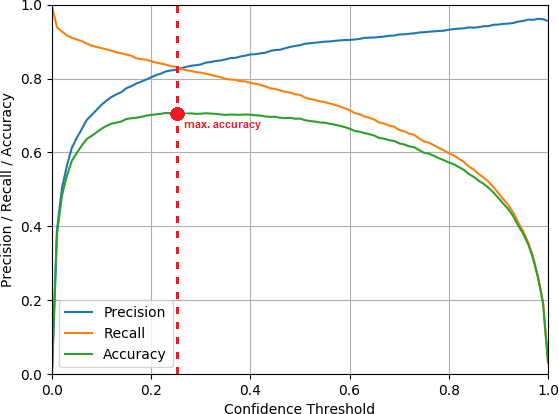}}
\caption{Graphs about sliding confidence threshold in text classification for the base leaflets dataset.}
\label{fig:results_base_th} 
\end{figure}

\vspace{0.2cm}
It must be remarked that a confidence threshold of $0.25$ is used for the \mbox{multi-label} text classification model of our proposal. This confidence represents the probabilities of having a correct prediction for a class, so the confidence threshold is used to filter predictions with low probabilities. The specific confidence threshold value is obtained by maximizing the accuracy values over threshold iterations from $0.00$ to $1.00$, as can be seen in the graph presented in Fig.~\ref{fig:results_base_th}~(b). To make fair comparisons, we also set the confidence threshold for the maximized accuracy value of the standard baseline approach, which is 0.40. The threshold iteration graph for the baseline method is shown in Fig.~\ref{fig:results_base_th}~(a).

\begin{figure}[!ht]
\centering
\subfigure[]                                     
{\includegraphics[width=0.325\textwidth]{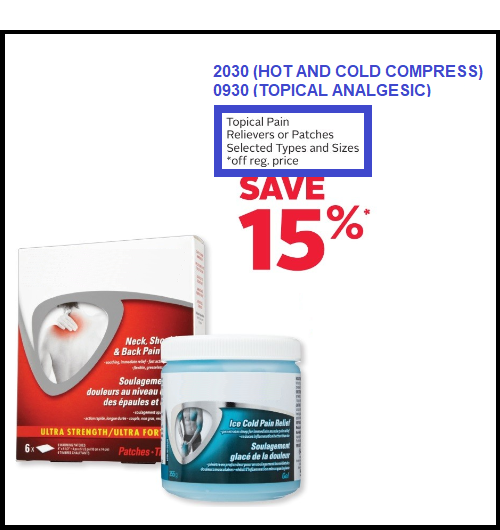}}
\subfigure[]                                     
{\includegraphics[width=0.325\textwidth]{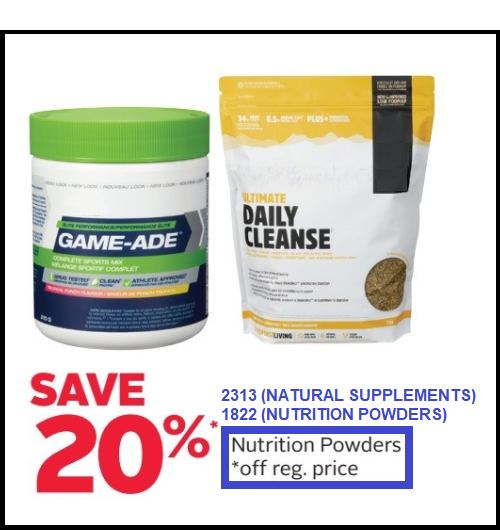}}
\subfigure[]                                     
{\includegraphics[width=0.325\textwidth]{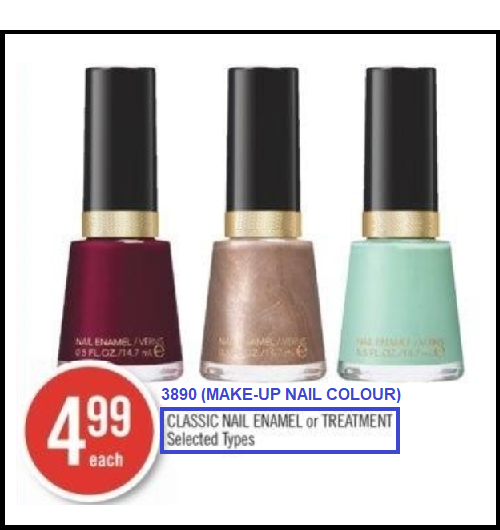}}
\subfigure[]                                     
{\includegraphics[width=0.325\textwidth]{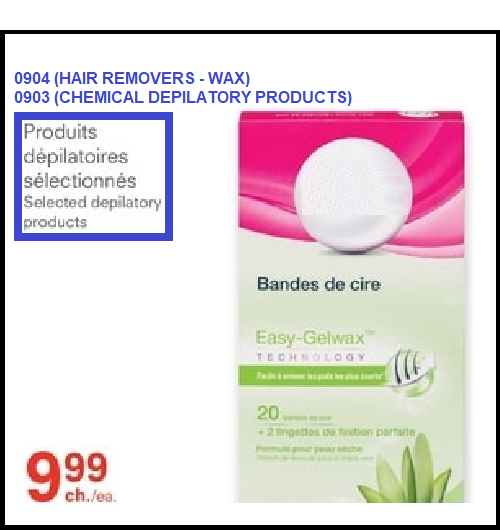}}
\subfigure[]                                     
{\includegraphics[width=0.325\textwidth]{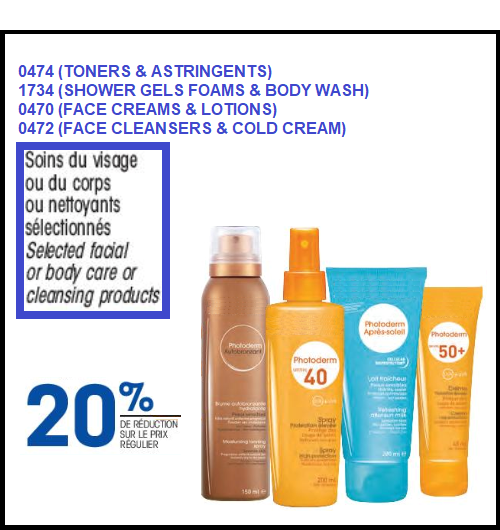}}
\subfigure[]                                     
{\includegraphics[width=0.325\textwidth]{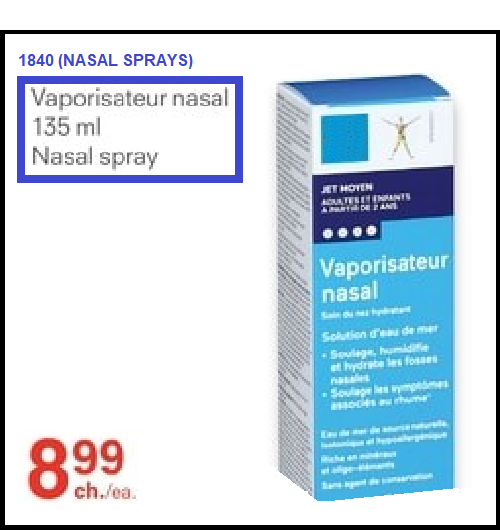}}
\caption{Qualitative results about leaflets examples and their corresponding predictions.}
\label{fig:qualitative_results} 
\end{figure}

As a final insight about results, we trained our models using the extended dataset to check out how it generalizes to more retailers and languages. The obtained results can be seen in Table~\ref{table:results_extended}, where the models trained in the extended dataset are having a better performance in test. Moreover, in Fig.~\ref{fig:qualitative_results} we depict some qualitative results about some leaflets and their corresponding predictions. According to these results, it seems that the embeddings for the text classifier are able to generalize categorization to new languages. The reported accuracies must be understood taking into account the \mbox{long-tail} problems of the dataset exposed in Fig.~\ref{fig:datasets_distribution}, so the classes with less training samples are more difficult to predict.
\begin{table}[!ht]
\begin{center}
\begin{tabular}{|l|c|c|c|}
\hline
\textbf{Dataset}  & \textbf{Precision} & \textbf{Recall} & \textbf{Accuracy} \\ \hline
Base     & 0.86               & 0.81            & 0.72              \\ \hline
\textbf{Extended} & \textbf{0.87}               & \textbf{0.86}            & \textbf{0.76}              \\ \hline
\end{tabular}
\end{center}
\caption{Comparison of results for our method in the base and extended leaflets datasets.}
\label{table:results_extended}
\end{table}

\section{Conclusions}
\label{sec:conclusions}

Along this paper, we have presented for the first time in the e-commerce research community (to the best of our knowledge) the problem of automated product coding for digital leaflets. In particular, we have addressed the problem of product classification for each promotion using image detection and \mbox{multi-label} text classification techniques. This schema provides a final proposal in the intersection between CV and NLP domains. Experimental results show that the described approach consistently outperforms a standard baseline in all the evaluated scenarios. 

Future research includes expanding the multi-label classification of each promotion to knowledge extraction of different attributes, such as brand and product names, quantities, volumes, price or discounts. The final goal of this research line is to extract all the possible information contained in digital leaflets in order to fully understand their whole context.
 
We believe that the automated product coding in digital leaflets is at an early research stage but yet it is a very interesting approach in the future of e-commerce. Then, this paper has contributed the initial milestones for the dissemination and enhancement of this research topic across the e-commerce research community.

\bibliographystyle{coling}
\bibliography{coling2020}

\end{document}